\PassOptionsToPackage{table}{xcolor}
\documentclass{article}

\IfFileExists{microtype.sty}{\usepackage{microtype}}{}
\usepackage{graphicx}
\usepackage{subcaption}
\usepackage{booktabs}
\usepackage{hyperref}

\usepackage[preprint]{icml2026}

\usepackage{caption}
\usepackage{amsmath}
\usepackage{amssymb}
\usepackage{array}
\usepackage{multirow}
\usepackage{makecell}
\usepackage{threeparttable}
\usepackage{pifont}
\usepackage[capitalize,noabbrev]{cleveref}
\newcommand{\method}{\textsc{PhysMRV}}

\newcommand{\best}[1]{\textbf{#1}}
\newcommand{\secondbest}[1]{\underline{#1}}

\icmltitlerunning{PhysMRV: Physical Memory Retrieval and Verification}

\begin{document}

\onecolumn

\icmltitle{PhysMRV: Physical Memory Retrieval and Verification\\for Physics Plausibility Reasoning}

\begin{icmlauthorlist}
  \icmlauthor{Wenyuan Wang}{ntu,rutgers}
  \icmlauthor{Lianyu Hu}{ntu}
  % \icmlauthor{Minghao Fu}{ucsd}
  \icmlauthor{Hao Wang}{rutgers,uiuc}
  \icmlauthor{Yang Liu}{ntu}
\end{icmlauthorlist}

\icmlaffiliation{ntu}{Nanyang Technological University}
\icmlaffiliation{rutgers}{Rutgers University}
\icmlaffiliation{uiuc}{University of Illinois Urbana-Champaign}
% \icmlaffiliation{ucsd}{University of California San Diego}

\icmlcorrespondingauthor{Wenyuan Wang}{ww462@scarletmail.rutgers.edu}
\icmlkeywords{Physical Reasoning, Visual-Language Models, Retrieval-Augmented Reasoning}

\vskip 0.3in

\printAffiliationsAndNotice{}

\begin{abstract}
Video-language models (VLMs) have achieved remarkable performance on video understanding and visual question answering, yet they remain unreliable in reasoning about physical plausibility, where understanding object interactions, causal dynamics, and fundamental physical principles is essential. This limitation is particularly evident on challenging physical reasoning benchmarks, revealing a persistent gap in physical commonsense reasoning.
To address this challenge, we propose \method{}, a training-free physical memory and verification framework for physical plausibility reasoning. Unlike retrieval-augmented VLMs that retrieve semantically similar videos as additional context, \method{} transforms training videos into a Hierarchical Memory Bank of structured physical knowledge comprising three complementary levels: scene descriptions capturing visual context, physical-event graphs modeling object interactions and causal structure, and physics-rule summaries distilling reusable physical principles and cues. During inference, \method{} retrieves physically relevant memories and leverages their structured physical evidence to guide a frozen VLM in verifying physical plausibility, requiring neither fine-tuning nor parameter updates.
We evaluate \method{} on three challenging physical reasoning benchmarks, ImplausiBench, IntPhys2, and GRASP Level 2, across multiple state-of-the-art VLMs. Experimental results demonstrate consistent improvements over direct prompting across diverse VLMs and evaluation benchmarks, showing that structured physical memories provide an effective and scalable means of enhancing physical plausibility reasoning without additional training.

\end{abstract}

\section{Introduction}

Recent VLMs have achieved impressive progress in video understanding, demonstrating strong performance in perceptual tasks and reasoning abilities, including spatial reasoning and temporal reasoning.\citep{bai2025qwen3vl,nvidia_cosmos_reason2_8b, wang2025internvl3_5,LLaVA-OneVision-2}. However, successful deployment in real-world environments requires capabilities beyond recognizing objects, actions, and scenes. A model also needs to reason about the physical world and determine whether an observed event is physically plausible. For example, it should recognize whether an unsupported object can remain suspended in midair, whether an object should continue to exist after becoming occluded, and whether a collision produces a physically consistent outcome. Such judgments require physical commonsense reasoning about object permanence, causal interactions, and the fundamental principles governing real-world dynamics. This makes physical plausibility reasoning fundamentally different from conventional video understanding: a model may correctly identify visible entities and actions while still failing to detect violations of basic physical laws, because such judgments cannot be directly captured by object-level semantics and instead require an understanding of real-world physical constraints.

\begin{figure}[!t]
    \centering
    \includegraphics[width=0.92\linewidth]{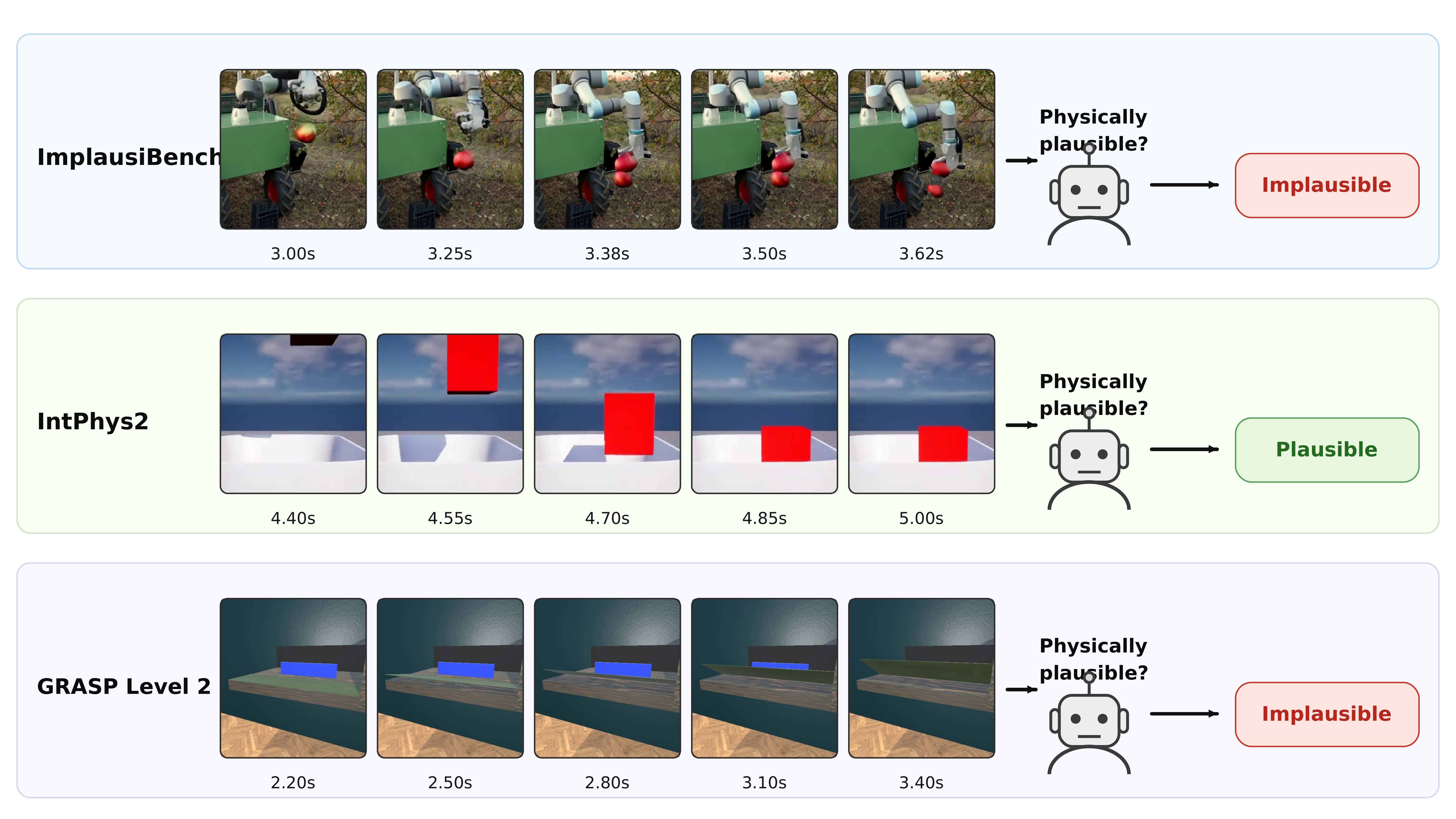}
    \caption{Examples of physical plausibility reasoning across the evaluation
  benchmarks. Given short video evidence, a model must determine whether the observed
  event is physically plausible rather than relying only on object or action
  recognition.}
    \label{fig:problem_illustration}
\end{figure}

Prior benchmarks have begun to systematically examine this capability. ImplausiBench focuses on detecting physical violations from visual evidence \citep{motamed2025travl}, IntPhys2 evaluates intuitive physics through violation-of-expectation scenarios involving object permanence, continuity, and solidity \citep{bordes2025intphys2}, and GRASP studies physical reasoning in grounded and interactive settings \citep{jassim2023grasp}.   Figure~\ref{fig:problem_illustration} shows representative examples from
  these benchmarks. Results on these benchmarks show that current multimodal models still fall substantially below human-level physical understanding, suggesting that VLMs lack a reliable mechanism for verifying whether observed evidence satisfies physical constraints.

This limitation is also consistent with the observation in InPhyRe \citep{sreekumar2026inphyrediscoverslargemultimodal}, which suggests that physical reasoning relies on identifying violations of latent physical constraints that are often invariant to surface-level visual similarity and thus cannot be reliably captured through retrieval signals based on semantic similarity.
A key reason is that current VLMs are strongly shaped by \emph{physical plausibility priors}. Since large-scale video-language pretraining data predominantly describe physically valid scenes, models tend to prefer physically reasonable interpretations even when the observed evidence suggests otherwise. As a result, their predictions are often driven by learned expectations about how the world typically behaves, rather than explicit verification of whether the current observation satisfies the relevant physical constraints. Effective physical reasoning therefore requires more than knowing what is physically likely; it requires verifying whether the observed visual evidence actually satisfies the relevant constraints.
Existing approaches have provided useful foundations. Prompting and context-construction methods, such as Physics Context Builders \citep{balazadeh2025pcb}, provide structured scene descriptions for VLM reasoning. Training-based approaches, such as TRAVL \citep{motamed2025travl}, improve physical-implausibility recognition through supervision training, but require updating model parameters. Existing retrieval-based approaches provide a non-parametric way to enhance physical reasoning by incorporating external memories at inference time \citep{luo2025videorag,zhong2024memorybank,ma2026brainmem,li2026physmem}. However, they are typically driven by semantic or task-level similarity, which is insufficient for physical plausibility reasoning. Visually or linguistically similar events may differ in underlying physical constraints such as support, contact, and causality. As a result, retrieval based solely on surface-level similarity may fail to distinguish physically distinct but semantically similar events, leading to unreliable plausibility verification. This suggests that retrieval should be conditioned on physical event structure rather than surface-level similarity.

To address this challenge, we propose \method{}, a non-parametric and plug-and-play framework for physical plausibility reasoning in video-language models. The core idea is to transform training videos into a structured \textit{Physical Memory Bank} that serves as a repository of explicit physical memories for inference. Each memory instance corresponds to a QA pair represented by hierarchical semantic-level scene caption, event-level physical graphs, and rule-level physical rules with associated positive and negative cues. At inference time, \method{} consists of two tightly coupled stages: \textbf{physics-aware retrieval} and \textbf{evidence-centered verification}. First, a coarse-to-fine retrieval strategy identifies physically analogous memories by matching semantic context and subsequently refining candidates using event-graph similarity. The retrieved memories are then treated as explicit physical evidence for verification, where a frozen VLM compares the observed event graph against retrieved rules and cues to determine whether the current event is physically plausible. Unlike conventional retrieval-augmented approaches that primarily retrieve textual context to enrich prompting, \method{} explicitly separates retrieval from verification, enabling frozen VLMs to validate physical plausibility through structured physical memories.

We evaluate \method{} on ImplausiBench, IntPhys2, and GRASP Level 2 across multiple vision-language model backbones, including Qwen3-VL, Cosmos-Reason2, InternVL3.5, and LLaVA-OneVision-2. Across all settings, \method{} consistently improves physical plausibility reasoning performance without any additional training or parameter updates to the backbone models.

Beyond overall performance gains, we conduct a series of analyses to understand where these improvements come from. Our results highlight the importance of structure-aware retrieval over purely semantic matching, the effectiveness of event graph-based alignment for identifying physically analogous cases, and the role of prompt-level grounding signals in enabling the frozen model to better leverage retrieved physical memories. Together, these findings suggest that explicitly structured physical memory can serve as an effective external mechanism for enhancing physical reasoning in frozen VLMs.

In summary, our contributions are:

\begin{itemize}

\item[\ding{182}] \textbf{\textit{Hierarchical Physical Memory Bank.}}
We construct a structured Physical Memory Bank from video QA training data, where each physical memory is represented by hierarchical semantic context, physical event graphs, and rule-level physical abstractions grounded in a unified taxonomy of physical principles.

\item[\ding{183}] \textbf{\textit{Physics-Aware Coarse-to-Fine Retrieval.}}
We propose a coarse-to-fine retrieval strategy that first recalls candidate physical memories using semantic context and then reranks them through physical event graph alignment, enabling frozen VLMs to retrieve physically analogous memories that support reliable physical plausibility verification.

\item[\ding{184}] \textbf{\textit{Training-free Physical Reasoning Framework.}}
We introduce \method{}, a plug-and-play framework that enhances frozen VLMs for physical plausibility reasoning by retrieving structured physical memories and using them as explicit evidence for physical plausibility verification, without any parameter updates or additional training.

\end{itemize}

\begin{figure}[!t]
    \centering
    \includegraphics[width=0.92\linewidth]{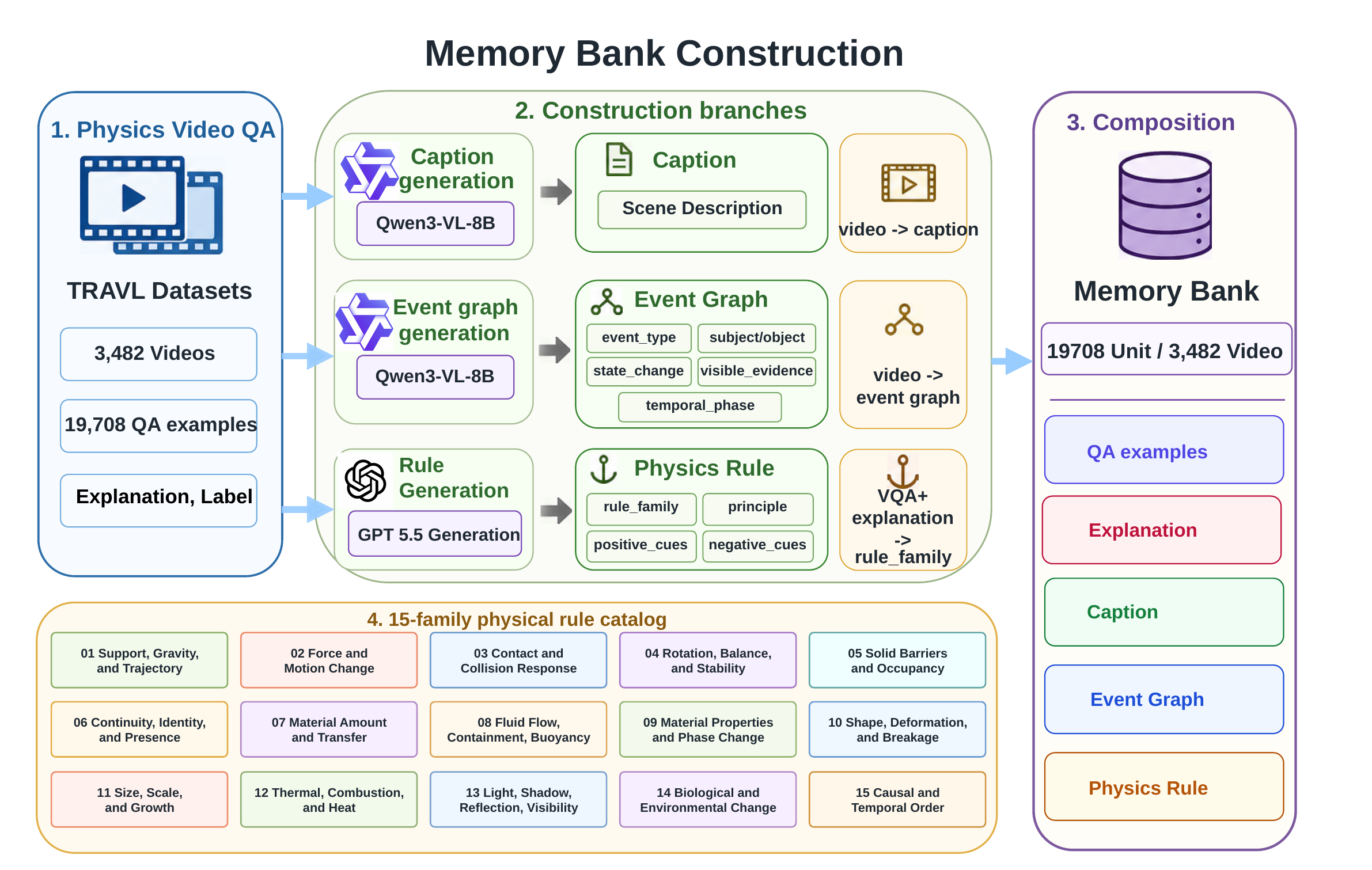}
    \caption{Structure of the Physics Memory Bank. Each training video contributes shared caption and physical-event rows, while question-answer anchor rows store training questions, training labels, explanations, rule-family assignments, principles, and cue lists. Dense recall and physical reranking select training-only anchors for the verifier.}
    \label{fig:memorybank-structure}
\end{figure}

\section{Related Work}

\subsection{Video Understanding}
VLMs aim to transform visual streams into representations suitable for question answering, event understanding, and reasoning. Despite strong progress in multimodal perception, existing models often struggle to preserve fine-grained temporal dependencies over long videos, especially when reasoning requires tracking interactions among multiple objects across time.

Recent work addresses this challenge from different perspectives. One line improves multimodal backbones and video interfaces to better capture long-range temporal structure, as exemplified by Qwen3-VL \citep{bai2025qwen3vl}. Another line focuses on efficient long-video comprehension through structured representations. LLoVi converts long videos into language-centered memory for question answering \citep{zhang2024llovi}, VideoTree organizes video content into hierarchical structures for LLM-based reasoning \citep{wang2024videotree}, and Video-RAG retrieves aligned visual and textual evidence for long-context video understanding \citep{luo2025videorag}.

Despite these advances, existing systems still largely rely on object- and action-level abstractions and often lack explicit modeling of fine-grained physical interaction structure, such as contact, support relations, and motion continuity, which are crucial for determining physical plausibility.

\subsection{Memory and Retrieval}
Retrieval-augmented methods enhance frozen models by incorporating external demonstrations, documents, or past experiences. In in-context learning, prior work shows that the choice of retrieved demonstrations significantly influences model behavior, motivating learned retrieval mechanisms that select examples tailored to a given query \citep{liu2022makes,rubin2022learning}.

Beyond short-term in-context retrieval, long-term memory systems maintain reusable experiences that can be revisited during inference. Memory Bank stores and retrieves persistent conversational memories for LLMs \citep{zhong2024memorybank}, BrainMem organizes embodied-agent experience into structured memory for planning and decision making \citep{ma2026brainmem}, and PhysMem introduces test-time memory for physical reasoning in embodied settings \citep{li2026physmem}.

These methods share a common paradigm of constructing searchable memory representations, retrieving relevant experiences, and injecting them into frozen models. However, most retrieval mechanisms are primarily optimized for semantic or task-level similarity, and may not explicitly preserve underlying physical interaction structure. As a result, retrieved examples can be content-relevant while still differing in fine-grained dynamics such as contact patterns, support relations, and state transitions.

This limitation is particularly critical for physical plausibility reasoning, which requires identifying whether observed events are consistent with underlying physical constraints rather than surface-level semantic similarity alone.

\subsection{Physical Reasoning}
Physical plausibility reasoning concerns an agent’s ability to infer physical states, interactions, and temporal transitions from observations. Recent work has extended its evaluation to VLMs through a diverse set of benchmarks. TRAVL and ImplausiBench evaluate whether models can distinguish physically plausible from implausible events in videos \citep{motamed2025travl}; IntPhys2 focuses on intuitive physics reasoning in synthetic environments \citep{bordes2025intphys2}; GRASP studies grounded physical understanding in situated settings \citep{jassim2023grasp}; and PhysBench provides a broader evaluation of physical-world understanding in multimodal models \citep{chow2025physbench}. Collectively, these benchmarks indicate that strong visual recognition and video understanding do not reliably transfer to physical plausibility reasoning, highlighting a persistent gap in modeling physical interactions.

Existing approaches to improving physical plausibility reasoning can be broadly categorized into two directions. The first direction specializes the reasoning model through task-specific training or supervision. For example, PhysGame constructs PhysInstruct and PhysDPO to train PhysVLM for detecting physical commonsense violations in gameplay videos \citep{cao2024physgameuncoveringphysicalcommonsense}, and TRAVL further improves performance through auxiliary supervision signals \citep{motamed2025travl}.

The second direction introduces auxiliary physical context while keeping the downstream model largely frozen. Physics Context Builders (PCBs), for instance, train a separate model on simulated physical scenes to generate physics-enriched context for downstream reasoning \citep{balazadeh2025pcb}.

While effective, these approaches typically rely on additional supervision or training, either for the reasoning model or for auxiliary context generation modules, which limits their applicability to frozen or proprietary VLMs and makes adaptation to new settings costly.

In contrast, we construct a physical experience memory from available data and perform retrieval and structured verification over retrieved cases entirely at inference time. This enables improved physical plausibility reasoning for frozen VLMs without requiring additional training or modification of the underlying model.

\section{Method}
\begin{figure}[!t]
    \centering
    \includegraphics[width=1.0\linewidth]{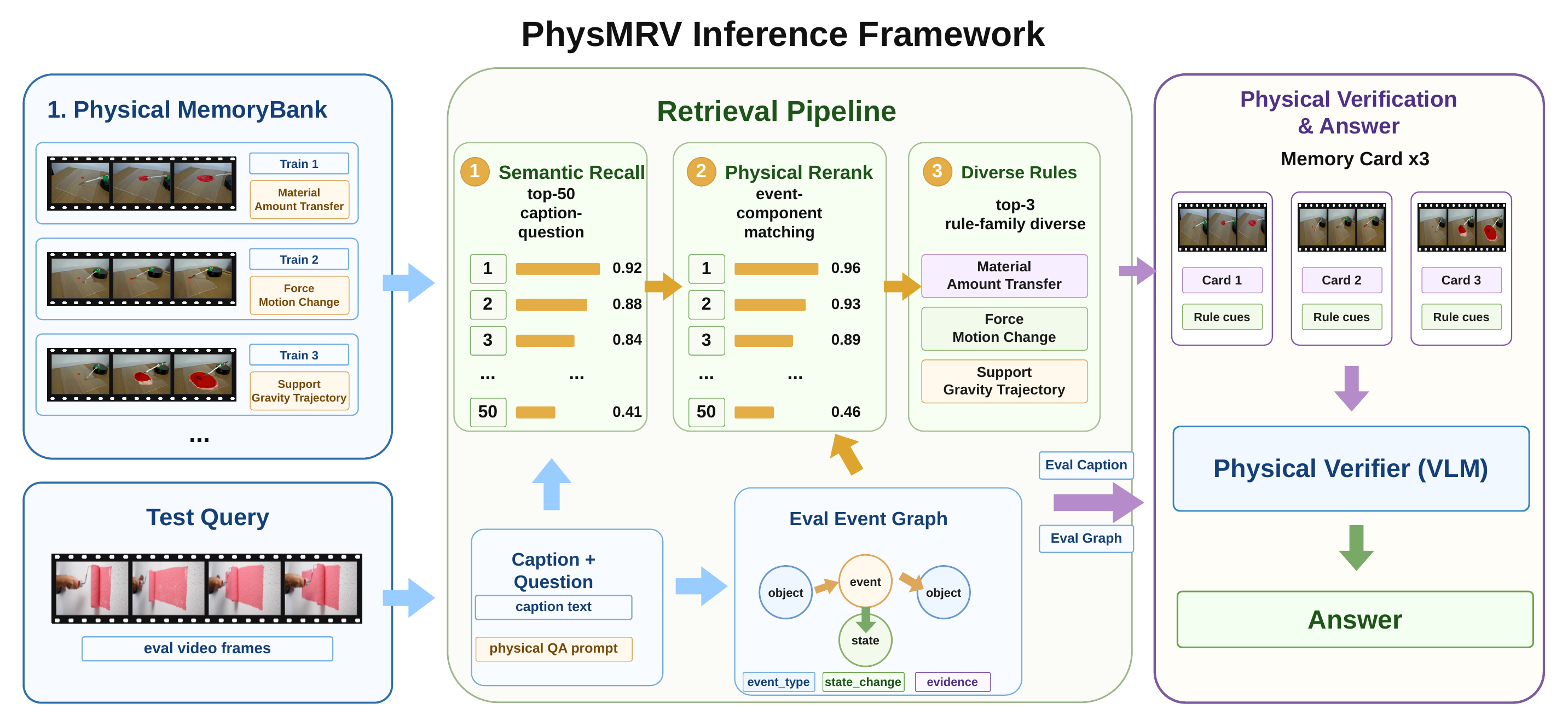}
 \caption{Overview of \method{}. Given a test video and question, \method{} retrieves physically relevant training precedents from Physical Memory Bank via hierarchical retrieval: semantic similarity identifies candidate scenes, and physical-event matching selects the most event-relevant memories across diverse physical phenomena. A frozen VLM then uses the video's extracted scene description and event graph to verify plausibility against the retrieved physical rules and diagnostic cues.}

    \label{fig:method}
\end{figure}
\vspace{-3mm}

\noindent \textbf{Overview.}

\method{} is a training-free inference-time framework for physical plausibility reasoning in frozen video-language models. It reformulates physical plausibility reasoning as a verification problem against retrieved physically similar training instances, which serve as explicit physical evidence for the current observation. Instead of directly inferring answers from video inputs, \method{} conditions reasoning on a structured set of retrieved physically similar memories, which provide explicit physical cues for verification.

The framework consists of three stages: (i) hierarchical Physical Memory Bank construction from TRAVL training videos, (ii) coarse-to-fine retrieval of physically relevant memories from the Memory Bank, and (iii) rule-guided verification via structured prompting, where retrieved physical rules and cues are explicitly used to guide the model’s plausibility judgment.

Figure~\ref{fig:memorybank-structure} illustrates the design of the Physical Memory Bank, and Figure~\ref{fig:method} presents the overall inference pipeline.

\subsection{Preliminaries}

We consider a physical plausibility reasoning problem over a benchmark set $\mathcal{B}$, which consists of ImplausiBench, IntPhys2, and GRASP Level~2. Each benchmark instance is defined as:
\begin{equation}
s_i = (v_i, q_i, \mathcal{A}_i, y_i),
\end{equation}
where $v_i$ is the input video, $q_i$ is the associated question, $\mathcal{A}_i$ denotes the candidate answer set, and $y_i \in \mathcal{A}_i$ is the ground-truth label. For multiple-choice benchmarks (e.g., ImplausiBench), $\mathcal{A}_i$ is explicitly provided, while for binary classification benchmarks (e.g., IntPhys2 and GRASP Level~2), $\mathcal{A}_i$ is mapped to a binary label space.

The task is to predict the correct answer given only the test-time input $(v_i, q_i)$:
\begin{equation}
\hat{y}_i = \arg\max_{a \in \mathcal{A}_i} P(a \mid v_i, q_i),
\end{equation}
without any training or parameter updates on $\mathcal{B}$.

\paragraph{Physical Memory Bank Construction.}
\method{} constructs an offline Physical Memory Bank $\mathcal{M}$ using training videos from TRAVL. Each memory entry $m_j \in \mathcal{M}$ is defined as:
\begin{equation}
m_j = (cap_j^{\mathrm{train}}, g_j^{\mathrm{train}}, q_j^{\mathrm{train}}, e_j, r_j, C_j^+, C_j^-),
\end{equation}
where $cap_j^{\mathrm{train}}$ is the generated video caption, $g_j^{\mathrm{train}}$ denotes the extracted physical event graph, $q_j^{\mathrm{train}}$ is the associated training question, $e_j$ is the dataset-provided explanation, $r_j$ is the corresponding physical rule, and $C_j^+$ and $C_j^-$ represent positive and negative physical cue sets, respectively.

\paragraph{Test-Time Setting.}
At inference time, given a test instance $(v_i, q_i)$, \method{} retrieves a relevant subset $R_i \subset \mathcal{M}$ based on semantic similarity in the representation space of the frozen video-language model $F_\theta$. The retrieved evidence is then organized into a structured verification prompt $P_i$, which provides explicit physical context for reasoning.

Given $(v_i, q_i, P_i)$, a frozen video-language model $F_\theta$ produces the final prediction:
\begin{equation}
\hat{y}_i = F_\theta(v_i, q_i, P_i),
\end{equation}
without any updates to model parameters $\theta$.

\subsection{Hierarchical Physical Memory Bank}

Physical plausibility reasoning depends on both observed dynamics and the physical regularities behind them. We therefore construct a Hierarchical Physical Memory Bank from 3,482 TRAVL training videos with 2--6 QA pairs per video, yielding 19,708 VQA examples. The Physical Memory Bank organizes each video into three complementary abstraction levels:
\begin{itemize}
    \item \textbf{Caption Layer}, which stores coarse-grained global scene descriptions;
    \item \textbf{Physical-Event Layer}, which stores fine-grained event graphs for object interactions and state transitions;
    \item \textbf{Rule Layer}, which stores the physical principle, and diagnostic cues.
\end{itemize}
For each video, the caption and physical-event graph are generated once and shared by all associated QA memories. Each QA memory is linked to one of fifteen rule families covering support, force, collision, continuity, fluid, thermal, optical, and causal phenomena, enabling semantic retrieval through text and physical matching through structured event evidence.

This design separates \textit{what is seen} (caption), \textit{what happens} (event graph), and \textit{why it is valid or invalid} (physical rule).

Physical-event graphs are constructed by prompting Qwen3-VL-8B to extract objects, state transitions, temporal phases, and causal relations as a structured event list. Training graphs $g_j^{\mathrm{train}}$ are built offline during Memory Bank construction; the same pipeline generates evaluation graphs $g_i^{\mathrm{eval}} = \Gamma(v_i, q_i)$ at inference time from the target video.

\subsection{Physical Memory Retrieval and Verification}

Given a query video and question, retrieval identifies Physical Memory Bank samples that are both semantically related and physically relevant.

\paragraph{Scene-Level Semantic Retrieval.}

The first stage favors recall by using caption-question representations rather than questions alone: the caption describes the observed scene, while the question identifies the physical aspect to judge. For an evaluation sample, we form $\tau_i^{\mathrm{eval}}=\mathrm{cap}_i^{\mathrm{eval}}\oplus q_i^{\mathrm{eval}}$ and represent each training memory analogously as $\tau_j^{\mathrm{train}}=\mathrm{cap}_j^{\mathrm{train}}\oplus q_j^{\mathrm{train}}$.
We encode these texts with Qwen3-Embedding-0.6B, compare normalized embeddings by cosine similarity, and keep the top-50 candidates as $\mathcal{C}_i^{50}$.
This broad pool is then refined by physical-event reranking.

\paragraph{Physical-Event Reranking.}

Semantic similarity does not necessarily imply physical similarity, so
\method{} reranks $\mathcal{C}_i^{50}$ according to structured
physical-event graph similarity. Let $E_i^{\mathrm{eval}}$ and
$E_j^{\mathrm{train}}$ denote the event sets extracted from the current
graph $g_i^{\mathrm{eval}}=\Gamma(v_i,q_i)$ and a training candidate graph
$g_j^{\mathrm{train}}$. Each event is decomposed into the ordered tuple
$\mathcal{P}=(\mathrm{event\_type},\mathrm{objects},\mathrm{state},
\mathrm{evidence},\mathrm{phase})$.

For component $p\in\mathcal{P}$, let $\mathbf{u}_p(e)$ be the
$\ell_2$-normalized embedding of the textual representation of component
$p$ in event $e$. The physical match between an evaluation event $e$ and
a training event $e'$ is
\begin{equation}
M(e,e') = \sum_{p\in\mathcal{P}} \alpha_p\,
\mathbf{u}_p(e)^\top \mathbf{u}_p(e').
\label{eq:event-match}
\end{equation}
Here we set each component-specific weight $\alpha_p$ to 0.2. To measure how well a training memory covers the current video's
physical events, we define the graph-level score as
\begin{equation}
S_{\mathrm{graph}}(g_i^{\mathrm{eval}},g_j^{\mathrm{train}})
= \frac{1}{|E_i^{\mathrm{eval}}|}
\sum_{e\in E_i^{\mathrm{eval}}}
\max_{e'\in E_j^{\mathrm{train}}} M(e,e').
\label{eq:physical-score}
\end{equation}
Candidates are reranked by $S_{\mathrm{graph}}$. To encourage diverse
physical precedents, duplicate videos are removed and the final top-3
anchors $R_i$ are greedily selected such that each belongs to a
different rule family.

\subsection{Rule-Cue Verification}

After retrieval, the selected memories are rendered as compact verification criteria rather than answer templates.

For each selected memory $m_j\in R_i$, we expose a compact verification record by retaining only the fields required for reasoning:
\begin{equation}
V_j=
(q_j^{\mathrm{train}},
e_j,
r_j,
C_j^+,
C_j^-).
\end{equation}

These cues turn retrieved precedents into checks over the current video: positive cues describe physically consistent evidence, while negative cues describe typical violations involving contact, support, continuity, trajectory, state change, or material behavior.

Let $\mathcal{V}_i$ collect the verification records for all retrieved memories  $\mathcal{V}_i = \{V_j \mid m_j \in R_i\}$. The final prompt is generated as
\begin{equation}
P_i =
\rho(q_i,\mathcal{A}_i,\mathrm{cap}_i^{\mathrm{eval}},g_i^{\mathrm{eval}},\mathcal{V}_i),
\end{equation}
so the prompt-visible context contains the current question, answer, eval caption, eval physical-event graph, and retrieved rule-cue records. The VLM is instructed to act as a visual verifier: it compares the current video and event graph against the retrieved principles and cue lists, then answers from current evidence. The final prediction is
\begin{equation}
\hat y_i =F_\theta(P_i),
\end{equation}
where the model is prompted to respond with a single token from $\mathcal{Y}_b$.

\section{Experiments}

\textbf{Datasets.}
We evaluate on three video benchmarks for physical plausibility reasoning.
ImplausiBench~\citep{motamed2025travl} contains 150 paired scenarios and 300 videos, with one plausible and one implausible video per scenario, and uses multiple-choice questions about the relevant physical outcome or violation.
IntPhys2 Main~\citep{bordes2025intphys2} contains 1,012 synthetic videos balanced between possible and impossible events, covering object permanence, immutability, spatio-temporal continuity, and solidity.
GRASP Level~2~\citep{jassim2023grasp} contains 4,096 situated-physics videos with concept-specific yes/no questions about whether the observed event is physically possible.

\noindent \textbf{Metrics.}
We report each benchmark with its task-native evaluation metric.
For ImplausiBench, the public release provides the multiple-choice annotations but without an official option-accuracy evaluation script. We therefore implement a lightweight scorer that parses the model's returned option letter (A--G) and computes exact-match accuracy against the benchmark MCQA target.
For IntPhys2, the official code includes multiple prompting/parsing variants. We use the official binary $1/0$ convention, where $1$ denotes physically possible/plausible behavior and $0$ denotes impossible/implausible behavior. We report balanced possible/impossible accuracy, i.e., the average accuracy over the possible and impossible subsets, to reduce sensitivity to class-prior bias. For GRASP, we follow the official binary evaluation script. Model outputs are reduced to the first valid yes/no answer and compared with the ground-truth label, where \textit{yes} indicates a physically plausible video and \textit{no} indicates an implausible one.

\noindent \textbf{Implementation Details.}
We use Qwen3-VL-8B-Instruct to generate captions and physical-event graphs. Rule-family labels, cues, and rule assignments are all generated using the GPT-5.5 API. Unless otherwise specified, all video inputs are processed with \texttt{video\_fps}=1.0. Caption generation uses deterministic decoding with temperature 0 and \texttt{max\_new\_tokens}=160, while physical-event graph extraction additionally uses \texttt{video\_max\_pixels}=151200 and \texttt{max\_new\_tokens}=1280. Final answer generation is performed with deterministic decoding (temperature 0), using \texttt{max\_new\_tokens}=256 for ImplausiBench and IntPhys2, and 32 for GRASP. Videos are decoded using \texttt{torchcodec}. All experiments are conducted on 4 H100 GPUs.

\subsection{Main Results}

\begin{table}[!t]
\centering
\tiny
\setlength{\tabcolsep}{1.4pt}
\caption{Main results on three physical-reasoning benchmarks. For each final VLM, we compare zero-shot direct prompting with component-level controls: evaluation captions, current-video physical graphs, train-memory retrieval, and the full \method{}. Training memories are drawn exclusively from TRAVL~\citep{motamed2025travl}. Best and second-best results for each backbone and benchmark are highlighted.}
\label{tab:main-focused}
\resizebox{\linewidth}{!}{%
\begin{tabular}{lccccc ccccc ccccc ccccc}
\toprule
\rowcolor{blue!10}
\textbf{Benchmark}
& \multicolumn{5}{c}{\textbf{Qwen3-VL}}
& \multicolumn{5}{c}{\textbf{Cosmos-Reason-2}}
& \multicolumn{5}{c}{\textbf{InternVL3.5-8B}}
& \multicolumn{5}{c}{\textbf{LLaVA-OneVision-2}} \\
\cmidrule(lr){2-6}
\cmidrule(lr){7-11}
\cmidrule(lr){12-16}
\cmidrule(lr){17-21}
\rowcolor{blue!10}
& Dir. & +Cap. & +Graph & Mem. & \method{}
& Dir. & +Cap. & +Graph & Mem. & \method{}
& Dir. & +Cap. & +Graph & Mem. & \method{}
& Dir. & +Cap. & +Graph & Mem. & \method{} \\
\midrule

ImplausiBench option acc.
& 0.5500 & 0.5567 & \secondbest{0.5800} & 0.5667 & \best{0.5900}
& 0.5333 & 0.5367 & \secondbest{0.5533} & 0.5267 & \best{0.5733}
& 0.5433 & \secondbest{0.5767} & 0.5600 & 0.5633 & \best{0.5833}
& 0.3400 & \best{0.3600} & 0.3533 & 0.3333 & \secondbest{0.3567} \\

IntPhys2
& 0.4990 & \secondbest{0.5217} & 0.5040 & 0.5040 & \best{0.5395}
& 0.5010 & 0.5208 & 0.5109 & \secondbest{0.5336} & \best{0.5366}
& 0.4990 & \secondbest{0.5217} & 0.4980 & 0.5198 & \best{0.5346}
& 0.4802 & \secondbest{0.5000} & 0.4990 & 0.4921 & \best{0.5020} \\

GRASP L2
& 0.4836 & \secondbest{0.5542} & 0.5227 & 0.5386 & \best{0.5798}
& 0.4998 & 0.5405 & 0.5308 & \secondbest{0.5527} & \best{0.5664}
& 0.4973 & \secondbest{0.5479} & 0.5029 & 0.5415 & \best{0.5715}
& 0.4885 & 0.5029 & 0.4922 & \secondbest{0.5061} & \best{0.5081} \\

\bottomrule
\end{tabular}%
}
\end{table}

Table~\ref{tab:main-focused} compares \method{} with zero-shot direct prompting and four component-level controls across four representative VLM backbones.
The controls isolate the contributions of evaluation captions, current-video physical graphs, memory retrieval only, and the complete PhysMRV framework.

\noindent\textbf{\method{} consistently improves physical plausibility reasoning across strong VLM backbones.}
Across Qwen3-VL, Cosmos-Reason-2, and InternVL3.5-8B, \method{} achieves the best performance on all three benchmarks.
Relative to direct prompting, it improves option accuracy by approximately 4 percentage points on ImplausiBench and IntPhys2, while yielding substantially larger gains of 6--10 percentage points on GRASP Level~2.
The largest improvement is achieved by Qwen3-VL on GRASP Level~2, where accuracy increases from 0.4836 to 0.5798 ($+9.6$ pp).

\noindent\textbf{Verification provides complementary benefits beyond retrieval or context augmentation alone.}
Neither evaluation-caption augmentation, evaluation-graph augmentation, nor training memory retrieval without verification consistently matches the full framework across the three benchmarks.
While individual components occasionally improve on specific datasets, integrating retrieved physical memories with verification consistently produces the strongest overall performance, suggesting that both retrieving relevant physical memories and explicitly verification over captions and structured physical events are important for robust physical plausibility reasoning.

\noindent\textbf{The largest improvements are observed on GRASP Level~2.}
Compared with direct prompting, \method{} improves GRASP Level~2 by $+9.6$, $+6.7$, and $+7.4$ percentage points for Qwen3-VL, Cosmos-Reason-2, and InternVL3.5-8B, respectively.
GRASP Level~2 requires fine-grained discrimination between subtle physical violations and plausible events, a setting where retrieved physical memories provide particularly targeted evidence.

\noindent\textbf{LLaVA-OneVision-2 benefits only marginally from \method{}.}
Unlike the other backbones, LLaVA-OneVision-2 exhibits a near-degenerate prediction pattern, classifying most events as plausible regardless of the input.
Consequently, \method{} yields only marginal improvements on IntPhys2 and GRASP Level~2 and no improvement on ImplausiBench.
We include these results for completeness, highlighting that the effectiveness of retrieval-based reasoning ultimately depends on the backbone's ability to utilize the retrieved evidence.

\subsection{Ablation Studies}

We decompose \method{} into a three-level ablation to isolate the contribution of each component: (i) retrieval quality, (ii) scene-grounded verification, and (iii) multi-anchor aggregation. Table~\ref{tab:ablation-controls-current} reports all four variants on Qwen3-VL and InternVL3.5-8B across the three benchmarks.

\begin{table*}[t]
\centering
\tiny
\setlength{\tabcolsep}{1.0pt}
\caption{Ablation and retrieval strategy comparison. Main uses rule-family-diverse top-3 physical-event graph retrieval with current/evaluation graph and verifier context. Visual-CLIP and Semantic retrieval variants use the same verifier context as Main but replace graph-aware reranking with visual frame similarity and text-only semantic retrieval, respectively. Top-1 uses graph-aware retrieval with a single anchor. Top 3 memories only keeps query-matched memories but removes evaluation graph and verifier context. Random 3 uses randomly sampled train memories without verifier context. ImplausiBench cells use return-only option accuracy; IntPhys2 and GRASP cells use balanced accuracy.}
\label{tab:ablation-controls-current}
\resizebox{\linewidth}{!}{%
\begin{tabular}{lcccccc cccccc}
\toprule
\rowcolor{blue!10}
\textbf{Benchmark}
& \multicolumn{6}{c}{\textbf{Qwen3-VL}}
& \multicolumn{6}{c}{\textbf{InternVL3.5-8B}} \\
\cmidrule(lr){2-7}
\cmidrule(lr){8-13}
\rowcolor{blue!10}
& \makecell{\textbf{Random 3}\\\textbf{memories only}}
& \makecell{\textbf{Top 3}\\\textbf{memories only}}
& \makecell{\textbf{Semantic retrieval}\\\textbf{+ verifier}}
& \makecell{\textbf{Visual-CLIP}\\\textbf{+ verifier}}
& \makecell{\textbf{Top-1}\\\textbf{rule anchor}}
& \textbf{Main}
& \makecell{\textbf{Random 3}\\\textbf{memories only}}
& \makecell{\textbf{Top 3}\\\textbf{memories only}}
& \makecell{\textbf{Semantic retrieval}\\\textbf{+ verifier}}
& \makecell{\textbf{Visual-CLIP}\\\textbf{+ verifier}}
& \makecell{\textbf{Top-1}\\\textbf{rule anchor}}
& \textbf{Main} \\
\midrule

ImplausiBench option acc.
& 0.5533 & 0.5667 & 0.5767 & \secondbest{0.5833} & \best{0.5967} & 0.5900
& 0.5400 & 0.5633 & 0.5600 & 0.5600 & \secondbest{0.5733} & \best{0.5833} \\

IntPhys2
& 0.4980 & 0.5040 & 0.5128 & \secondbest{0.5237} & \secondbest{0.5237} & \best{0.5395}
& 0.5030 & 0.5198 & \best{0.5445} & 0.5158 & 0.5099 & \secondbest{0.5346} \\

GRASP L2
& 0.4988 & 0.5386 & 0.5635 & \secondbest{0.5657} & 0.5437 & \best{0.5798}
& 0.5156 & 0.5415 & 0.5430 & \secondbest{0.5542} & 0.5398 & \best{0.5715} \\

\bottomrule
\end{tabular}%
}
\end{table*}

\noindent\textbf{Memory retrieval is the primary driver of improvement.}
The most direct evidence for memory effectiveness comes from comparing \emph{Random~3 memories only} against \emph{Top 3 memories only}. Both variants provide exactly three in-context training examples without verification; the sole difference is whether those examples are drawn from random training samples or retrieved by query-aware rule-family matching. \emph{Top 3 memories only} consistently outperforms \emph{Random~3} on all benchmarks for both backbones.
The gap is largest on GRASP Level~2, where matched retrieval exceeds random retrieval by $+3.98$ points for Qwen3-VL and $+2.59$ points for InternVL3.5-8B. This rules out the hypothesis that physical-reasoning gains can be obtained by simply prepending arbitrary training demonstrations; the improvements depend on surfacing memories that are semantically and rule-family relevant to the current event.

\noindent\textbf{Scene-grounded verification adds further gains on top of matched memories.}
Adding the current-video scene description and physical-event graph as verifier context (\emph{Top-1 rule anchor} and \emph{Main}) provides a second layer of improvement over the matched-memory-only baseline (\emph{Top 3 memories only}). For Qwen3-VL, the full method raises ImplausiBench option accuracy from $0.5667$ to $0.5900$ and IntPhys2 from $0.5040$ to $0.5395$, confirming that grounding the verification step in the current-video scene description and event graph is complementary to retrieval quality.

\noindent\textbf{Rule-diverse top-3 aggregation improves robustness beyond single-anchor retrieval.}
Comparing \emph{Top-1 rule anchor} with \emph{Main} isolates the effect of aggregating multiple rule-diverse physical precedents. On ImplausiBench, the top-1 variant is slightly higher for Qwen3-VL (0.5967 vs. 0.5900), but this pattern does not hold for InternVL3.5-8B, where \method{} is higher (0.5833 vs. 0.5733). In contrast, \method{} consistently improves GRASP Level~2 for both backbones, increasing Qwen3-VL from 0.5437 to 0.5798 and InternVL3.5-8B from 0.5398 to 0.5715. On IntPhys2, \method{} is strongest for Qwen3-VL (0.5395), while InternVL3.5-8B shows a slightly higher score with semantic retrieval plus verification (0.5445). These results suggest that a single high-confidence precedent can be competitive on constrained multiple-choice option selection, but rule-diverse top-3 aggregation provides more reliable gains on the broader binary video plausibility benchmarks, especially GRASP Level~2.

\noindent\textbf{Physical-event graph retrieval outperforms semantic and visual retrieval alternatives.}
Table~\ref{tab:ablation-controls-current} also includes two alternative retrieval strategies that use the same verifier context as \method{} but replace graph-aware reranking with text-only semantic retrieval (\emph{Semantic retrieval + verifier}) and visual frame similarity (\emph{Visual-CLIP + verifier}).
Both alternatives improve over the random and matched-memory-only baselines, confirming that verifier context contributes meaningfully regardless of retrieval strategy.
However, neither achieves consistent gains across all three benchmarks for both backbones.
Visual-CLIP retrieval degrades on IntPhys2, where the synthetic video domain reduces the reliability of appearance-based matching, and tends to retrieve anchors with low rule-family diversity, limiting the coverage of physical evidence presented to the verifier.
Dense text retrieval is more robust to domain shift but cannot distinguish physically distinct events that share surface-level language, such as scenarios involving the same objects or actions under different physical constraints.
Physical-event graph reranking addresses both failure modes by matching on interaction structure rather than appearance or language, and the rule-family diversity constraint ensures that the retrieved anchor set covers complementary physical principles.
\method{} is the only configuration that outperforms direct prompting consistently across all three benchmarks for both backbones.

\subsection{Bias Analysis Across Benchmarks}

Aggregate accuracy on balanced benchmarks is insensitive to answer-polarity bias: a model that always responds ``yes'' can score 50\% without any physical understanding.
We therefore report per-class accuracy slices as a diagnostic to test whether observed gains reflect genuine physical reasoning or a shift in answer prior.
Table~\ref{tab:bias-slices} reports positive-class (plausible/yes) and negative-class (implausible/no) accuracies for Qwen3-VL and InternVL3.5-8B—the two strongest backbones across all three benchmarks—as representative probes of the failure mode.

\begin{table}[t]
\centering
\tiny
\caption{Class-conditional diagnostic accuracy across three benchmarks. P/Y and I/N denote accuracy on plausible/yes and implausible/no instances, respectively. On ImplausiBench, these slices measure binary plausibility bias, while the Metric column reports benchmark-native option accuracy over the specific physical-reason choices; hence it is not computed as the average of P/Y and I/N. On IntPhys2 and GRASP, the native task is binary, so Metric is the balanced average of the two slices.}
\label{tab:bias-slices}
\setlength{\tabcolsep}{4pt}
\resizebox{\textwidth}{!}{%
\begin{tabular}{llccccccccc}
\toprule
\rowcolor{blue!10}
\textbf{Model} & \textbf{Method} &
\multicolumn{3}{c}{\textbf{ImplausiBench}} &
\multicolumn{3}{c}{\textbf{IntPhys2}} &
\multicolumn{3}{c}{\textbf{GRASP L2}} \\
\rowcolor{blue!10}
 & &
\textbf{P/Y} & \textbf{I/N} & \textbf{Metric} &
\textbf{P/Y} & \textbf{I/N} & \textbf{Metric} &
\textbf{P/Y} & \textbf{I/N} & \textbf{Metric} \\
\midrule
\rowcolor{gray!6}
Qwen3-VL & Direct &
0.9867 & 0.2867 & 0.5500 &
0.7589 & 0.2391 & 0.4990 &
0.9453 & 0.0220 & 0.4836 \\
Qwen3-VL & \method{} &
0.9933 & 0.3267 & 0.5900 &
0.8103 & 0.2688 & 0.5395 &
0.6636 & 0.4961 & 0.5798 \\
\midrule
\rowcolor{gray!6}
InternVL3.5-8B & Direct &
0.9531 & 0.3209 & 0.5433 &
0.9862 & 0.0119 & 0.4990 &
0.9814 & 0.0132 & 0.4973 \\
InternVL3.5-8B & \method{} &
0.9933 & 0.2867 & 0.5833 &
0.8202 & 0.2490 & 0.5346 &
0.4990 & 0.6440 & 0.5715 \\
\bottomrule
\end{tabular}}
\end{table}

\noindent\textbf{Direct VLMs exhibit a severe positive-class prior across backbones.}
Both backbones achieve near-ceiling plausible-event accuracy in the direct baseline while implausible-event accuracy collapses.
On GRASP Level~2—a balanced benchmark of 4,096 yes/no questions—Qwen3-VL direct correctly identifies fewer than one in forty-five impossible events (0.0220), and InternVL3.5-8B fewer than one in seventy-five (0.0132).
On IntPhys2, InternVL3.5-8B is even more extreme: it achieves perfect possible-event recall (0.9862) while detecting essentially no impossible events (0.0119).
This pattern is consistent across backbones and suggests that frozen VLMs default to affirming physical normalcy when perceptual evidence is ambiguous.

\noindent\textbf{\method{} corrects the prior in both backbones, but with different magnitudes and directions.}
Negative-class accuracy improves on all three benchmarks for both backbones, ruling out the hypothesis that aggregate gains stem purely from reinforcing an existing positive prior.
However, the correction overshoots differently for each backbone.
On GRASP Level~2, Qwen3-VL negative accuracy rises from 0.0220 to 0.4961 while positive accuracy drops from 0.9453 to 0.6636, yielding a slightly positive-biased post-method profile.
InternVL3.5-8B undergoes a larger shift in the same direction: positive accuracy drops from 0.9814 to 0.4990 and negative rises from 0.0132 to 0.6440, flipping to a slightly negative-biased profile.
On IntPhys2, by contrast, both positive and negative slices improve for both backbones without a polarity trade-off, indicating that the method can sharpen both-class discrimination when the perceptual signal is sufficient.

\noindent\textbf{\method{} substantially mitigates answer-polarity bias across backbones.}
Although the two backbones converge to different operating points after applying \method{}, both exhibit a much more balanced prediction profile than their direct counterparts.
Qwen3-VL shifts from an extremely positive-biased distribution (0.9453/0.0220) to a substantially more balanced one (0.6636/0.4961), while InternVL3.5-8B moves from an even stronger positive prior (0.9814/0.0132) to a considerably more balanced profile (0.4990/0.6440).
These consistent improvements across both backbones indicate that \method{} effectively mitigates the severe positive-class bias exhibited by frozen VLMs, rather than simply reinforcing their original answer prior.

\subsection{Qualitative Analysis}

\begin{figure*}[!t]
\centering
\scriptsize
\setlength{\tabcolsep}{5pt}
\begin{tabular}{@{}p{0.48\textwidth}p{0.48\textwidth}@{}}
\textbf{GRASP: identity-continuity correction} &
\textbf{IntPhys2: implicit violation correction} \\
\includegraphics[width=\linewidth]{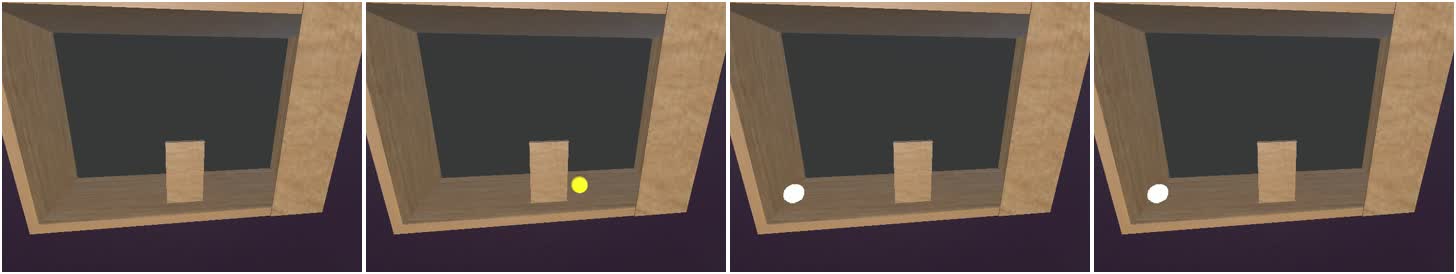} &
\includegraphics[width=\linewidth]{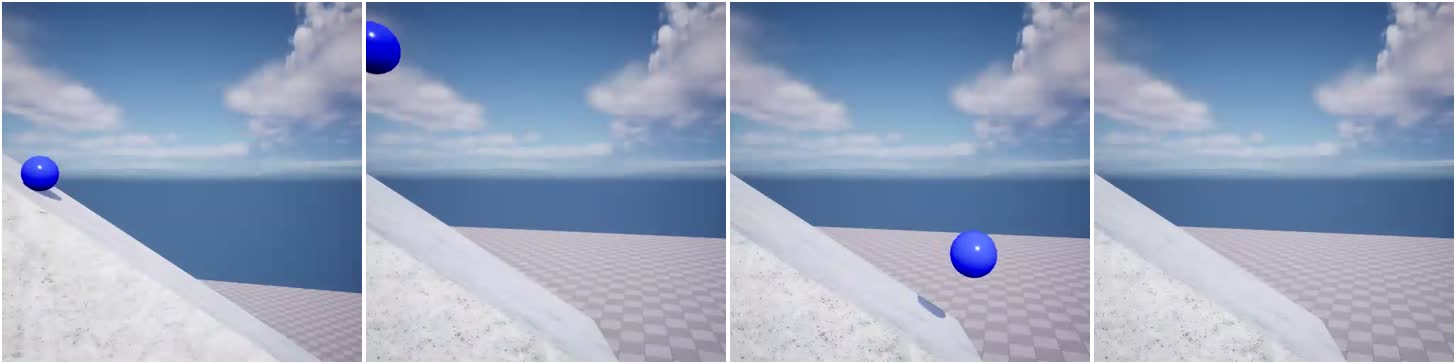} \\
\textbf{Video/question.} A yellow ball disappears behind a wooden block; a white ball later appears and rolls left. The task asks whether the outcome is plausible. &
\textbf{Video/question.} A blue sphere rolls down an incline, becomes airborne, lands, and disappears. The task asks whether the behavior is physically possible. \\
\textbf{Outputs.} Direct: yes; \method{}: no; GT: no. &
\textbf{Outputs.} Direct: possible; \method{}: impossible; GT: impossible. \\
\textbf{Retrieved criteria.} \emph{force\_motion\_change}, \emph{continuity\_identity\_presence}, and \emph{solid\_barrier\_occupancy}; cues include no visible force, vanishing, materializing, and passing through barriers. &
\textbf{Retrieved criteria.} \emph{continuity\_identity\_presence}, \emph{force\_motion\_change}, and \emph{contact\_collision\_response}; cues include unexplained appearance/disappearance and unsupported motion changes. \\
\textbf{Current graph fragment.} Yellow and white balls are separately tracked as moving objects behind the same block, exposing an identity change rather than ordinary occlusion. &
\textbf{Current graph fragment.} The sphere moves from the incline into the air and then vanishes after impact, exposing a continuity and force-motion violation. \\
\end{tabular}
\caption{Qualitative successful corrections. Left: a GRASP example where direct prompting follows the plausible/yes prior, while \method{} uses retrieved continuity and force-change criteria to answer the correct ``no.'' Right: an IntPhys2 example where retrieved continuity and motion cues help change a direct possible answer to the correct impossible answer.}
\label{fig:qualitative-cases}
\end{figure*}

Figure~\ref{fig:qualitative-cases} shows two examples where training only memories change an initially plausible answer into the correct implausible judgment.
In the GRASP case, direct prompting answers ``yes'' even though the video contains an identity-continuity violation: a yellow ball disappears behind a block and a white ball later appears.
The retrieved anchors expose force-change, continuity, and solid-barrier criteria, including negative cues such as objects that vanish, materialize, or move without visible cause.
With these criteria and the current event graph visible, \method{} answers the correct ``no.''

The IntPhys2 case shows the same mechanism on a different benchmark format.
Direct prompting marks the video as possible, while \method{} marks it impossible.
The current caption and event graph describe a blue sphere rolling down an incline, becoming airborne, and disappearing after impact.
Retrieved continuity and force-motion anchors provide the relevant checklist: objects should not appear or vanish without an explained transition, and motion changes should follow visible forces or contact.
In both cases, the retrieved rule families—\emph{continuity\_identity\_presence} and \emph{force\_motion\_change}—directly correspond to the violation type surfaced in the current-video graph, confirming that rule-family-aware retrieval selects memories governed by the same physical principles as the query event.
These cases make the aggregate prior-correction result concrete: \method{} does not merely add physics vocabulary, but brings matched training memories into the final decision context.

\section{Conclusion}

We introduced \method{}, a training-only physical precedent engine that enhances physical plausibility reasoning for frozen VLMs without any test-time model updates.
The method organizes TRAVL training examples into a hierarchical MemoryBank spanning semantic descriptions, physical-event components, and a compact catalog of physical rule families, and uses coarse-to-fine retrieval to supply rule-family-diverse case-level exemplars at inference time.
\method{} consistently improves three of four tested backbones across ImplausiBench, IntPhys2, and GRASP Level~2, with gains of up to 9.6 points on GRASP.
LLaVA-OneVision-2 improves on IntPhys2 and GRASP but not on ImplausiBench, suggesting that the benefit of retrieved physical precedents depends on backbone instruction-following capacity.

Three findings from our analysis sharpen the research picture.
First, case-level exemplar retrieval is the primary driver of improvement: randomly sampled in-context demonstrations are consistently weaker than rule-family-matched precedents, confirming that physical reasoning gains depend on surfacing semantically relevant training evidence rather than simply injecting in-context examples.
Second, direct VLMs exhibit a systematic positive-class prior that approaches near-ceiling on plausible events while collapsing on implausible events; \method{} substantially corrects this bias on GRASP but does not eliminate the calibration gap on IntPhys2.
Third, the magnitude and direction of polarity correction are backbone-dependent: Qwen3-VL and InternVL3.5-8B land on opposite sides of the balance point after correction, indicating that evidence integration is still mediated by backbone-specific priors rather than by per-instance physical reasoning.

Together, these results establish retrieved physical memory as an effective inference-time scaffold, while identifying calibrated per-instance physical verification—grounding decisions in evidence about specific object states, contacts, and trajectories observed in the current video—as the key open challenge for future physical plausibility systems.

\bibliographystyle{icml2026}
\bibliography{references}

\typeout{get arXiv to do 4 passes: Label(s) may have changed. Rerun}
\end{document}